\pdfoutput=1

\documentclass[conference, compsoc]{IEEEtran}

\usepackage{graphics}
\usepackage{epsfig} 
\usepackage{subfigure}
\usepackage{multirow}
\usepackage{hyperref}

\title{\LARGE \bf
Grasping the Inconspicuous
}
\author{Hrishikesh Gupta$^{*}$, Stefan Thalhammer$^{*}$, Markus Leitner, Markus Vincze
\thanks{*Equal contribution, All authors are with the Automation and Control Institute (ACIN), TU Wien, 1040 Vienna, Austria {\tt\small 
\{gupta, thalhammer, leitner, vincze\}@acin.tuwien.ac.at}%
}}

\begin{document}

\maketitle

\begin{abstract}

Transparent objects are common in day-to-day life and hence find many applications that require robot grasping. Many solutions toward object grasping exist for non-transparent objects. However, due to the unique visual properties of transparent objects, standard 3D sensors produce noisy or distorted measurements. Modern approaches tackle this problem by either refining the noisy depth measurements or using some intermediate representation of the depth. Towards this, we study deep learning 6D pose estimation from RGB images only for transparent object grasping. To train and test the suitability of RGB-based object pose estimation, we construct a dataset of RGB-only images with 6D pose annotations. The experiments demonstrate the effectiveness of RGB image space for grasping transparent objects.

\end{abstract}

\section{Introduction}

Object detection and pose estimation are two of the most fundamental problems in the field of robot vision, crucial for robotic object grasping and manipulation. Although robot object manipulation by means of pose estimation itself is a quite challenging problem, it still offers a good and wide range of solutions for manipulating opaque objects. Towards this considerable research has been devoted to robotic manipulation of objects using 3D data (e.g. RGB-D images, point clouds) ~\cite{ten2018using} ~\cite{zeng2018robotic}. However, many of these algorithms cannot be immediately applied to transparent objects~\cite{sajjan2020clear}. This is because, existing commercial depth sensors, such as projected light or time-of-flight sensors, assume that objects have Lambertian surfaces that can support diffuse reflection from the sensor. Depth sensing fails when these conditions do not hold, e.g., for transparent or shiny metallic objects. Transparent objects are a common part of everyday life, from reading glasses to plastic bottles – yet these unique visual and material properties make them difficult for machines to perceive and manipulate, especially with the mentioned sensors.

Our main premise for the experimental setup in this manuscript is that RGB images provide enough information for object pose estimation for transparent objects. Towards this, we proposed an experimental setup using a canister as a transparent object. Which is a sterile medical object often used in the medical field for temporary storing and processing of fluids. Hence, has a strong use-case for robot object grasping.
\begin{figure}[t]
      \centering
      \includegraphics[scale=0.33]{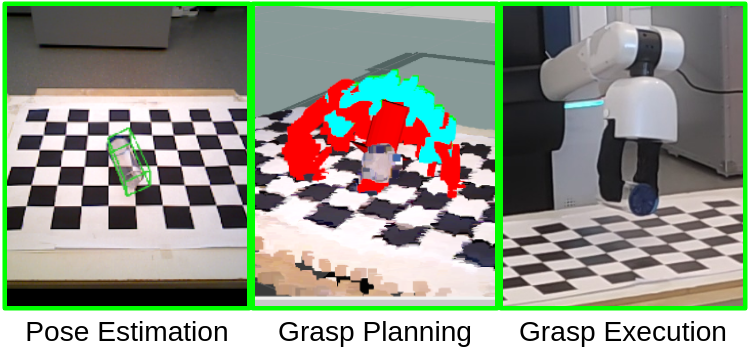}
      \caption{\textbf{Overview} Our method leverages RGB images for transparent object grasping. We first  perform pose estimation, grasp annotation, and planning, and as a final step execution of the grasp.}
      \label{Overview}
   \end{figure}

For evaluation of our assumption, we propose an experimental setup: 1) First, collect an RGB dataset for our transparent object. 2) Annotate it using a state-of-art 6D pose estimation annotation tool ~\cite{markus20193dsat}. 3) Use the state-of-art pose estimation method ~\cite{thalhammer2022cope} for pose estimation for our transparent object. 4) Grasping of the pose-estimated transparent object, to evaluate our assumption and the effectiveness of the RGB-only image space.
An overview of the grasping process is shown in Figure~\ref{Overview}.
We further qualitatively and quantitatively evaluate our experiments which prove the premise of our experimental setup. 

The following sections are organized as, section 3 describes our experimental setup. Where the pose estimation pipeline is discussed, along with the collection and annotation of the training data. In section 3 we also describe the grasping pipeline we use to evaluate our assumption for the usage of RGB only images. In Section 4 we present the experimental results of the transparent object grasping and discuss the results. Finally section 5 we talk in brief about our findings and possible future works.

\section{Related Work}

Recent work tackling transparent object grasping and manipulation lies at the intersection of object detection, segmentation, geometric reasoning, depth reconstruction, and boundary detection. 
Applying these challenging problems to transparent objects received increased attention lately.

Classical methods mostly rely on peculiarities of such objects, such as specular reflections and local characteristics of edges due to refraction ~\cite{mchenry2005finding}. ~\cite{fritz2009additive} used an additive model of latent representations to learn the appearance of transparent objects and remove the influence of background. These methods were made to perform localization of the objects and showed promising results in small experiments. Methods for transparent object segmentation started with the focus on formulating an energy function based on Light-Field linearity (LF-Linearity)~\cite{xu2015transcut} and occlusion detection from the 4D light-field image were optimized to generate the segmentation images. Recently, ~\cite{xie2020segmenting} introduce the Translab model for transparent object segmentation. They also introduced the first large-scale real-world transparent object segmentation dataset, termed Trans10K. It has 10K+ images. 
One of the most recent methods~\cite{kalra2020deep} combine polarization with deep learning and propose a polarized CNN for transparent object segmentation. Compared with previous methods, this still requires additional input data(Polarizing light-field) apart from the RGB only.

Methods such as TOM-Net~\cite{chen2018tom} addresses the problem of transparent object matting. And formulating the problem as a refractive flow estimation problem. They propose a multi-scale encoder-decoder network to generate a coarse input, and then a residual network refining it to a detailed matte.

When it comes to transparent object pose estimation, initial methods either leverage failure modes of depth sensors like Microsoft Kinect\footnote{\url{https://en.wikipedia.org/wiki/Kinect}} and estimate object pose with a known 3D shape model ~\cite{wang2012glass} ~\cite{lysenkov2013recognition}, or also structured light sensors ~\cite{phillips2016seeing}. Because of the requirement for prior object models or specific sensors, those approaches do not allow a simple scenario where the only sensor is a stereo camera, or a commercial camera taking a couple of pictures of a scene “in the wild”.

Many of the methods for transparent objects pose estimation rely on depth information.
Hence many recent methods try to complete the missing depth information for transparent objects. The most recent and relevant methods closest to our work provide depth completion from an RGB image with inaccurate depth information ~\cite{sajjan2020clear}~\cite{zhu2021rgb}.

Transparent objects have been previously studied in various computer vision applications, including object pose estimation ~\cite{klank2011transparent} ~\cite{lysenkov2013pose} ~\cite{lysenkov2013recognition}. Works on estimating transparent object pose and geometry might assume knowing the object 3D model ~\cite{lysenkov2013recognition} ~\cite{phillips2016seeing}. In ~\cite{lysenkov2013pose}~\cite{lysenkov2013recognition}, the pose of a rigid transparent object is estimated by 2D edge feature analysis. In ~\cite{guo2019transparent}, SIFT features are used to recognize the transparent object. However, low-level traditional features are not as discriminative as high-level deep features. The most recent method for Pose Estimation of the transparent objects introduced a keypoint-based feature~\cite{liu2020keypose} for pose estimation, trained on stereo images. But this requires manually choosing keypoints that should best describe the object pose with the addition of stereo images instead of RGB only.

All these methods heavily rely on additional input information besides RGB, for transparent object pose estimation. Hence requiring more complicated annotation processes such as in~\cite{liu2020keypose} and models. Hence we put forward our method requiring RGB only information for transparent object pose estimation and grasping. We show through our experiments, both qualitatively and quantitatively, that RGB images provide enough information for transparent object manipulation.

\section{Experimental Setup}

To demonstrate the potential of RGB for transparent object grasping, an experimental setup for pose estimation and manipulation of a transparent canister is created.
The canister, shown in Figure~\ref{Dataset} is a medical sterile object often used in the medical industry. In recent times due to growth in the automation of the medical sector, grasping such medical transparent objects canister is often one of the encountered hurdles in robotics.

\begin{figure}[t]
      \centering
      \includegraphics[scale=0.23]{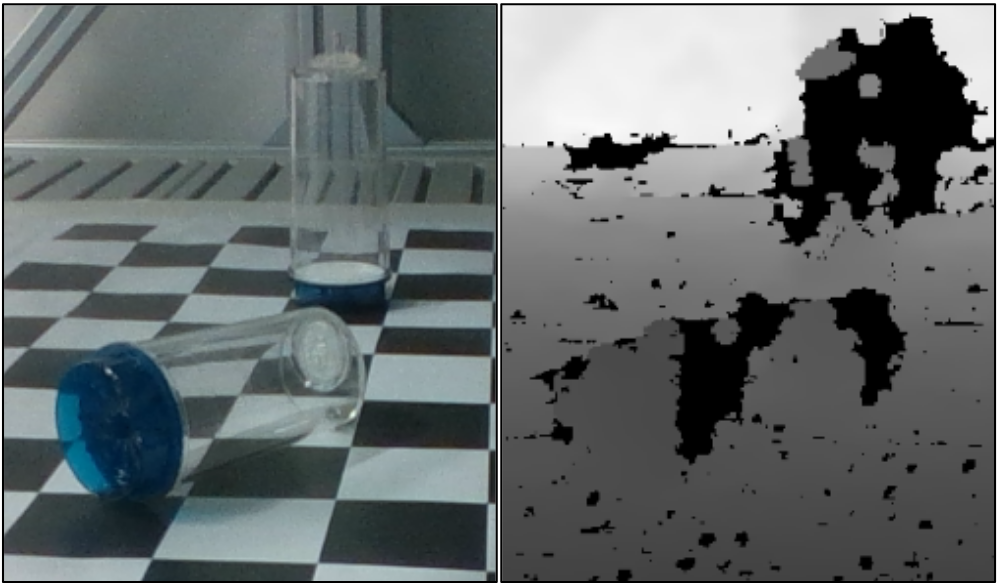}
      \caption{\textbf{Point cloud of the Canister} Missing depth information of the transparent canister due to its non-lambertian nature.}
      \label{PointcloudofCanister}
   \end{figure}

Towards our experimental setup, firstly we collected a dataset of the transparent object and annotate it.
A state-of-art pose estimation method~\cite{thalhammer2022cope} is trained on the collected dataset.
The trained estimator is deployed in a robotic setup, to estimate the canister's pose on a tabletop.
Based on this estimate the object is grasped.

\subsection{Pose Estimation}

Most robot grasping methods require object orientation and localization information in the scene. Hence, pose estimation is often a predecessor step for robot object grasping. Most transparent object pose estimation methods use some form of depth information. Since depth information for transparent objects produces degenerate solutions (see Figure~\ref{PointcloudofCanister})~\cite{sajjan2020clear}~\cite{zhu2021rgb}, most of these methods add an extra step of either depth refinement ~\cite{sajjan2020clear}~\cite{zhu2021rgb}, extracting depth information of the background for refinement~\cite{xu20206dof}, or use some other form of the intermediate representation of the depth like disparity maps~\cite{liu2020keypose} for pose estimation. 

Since the basic assumption of this work is that RGB images provide suitable information for transparent object grasping, we use a recent RGB-based object pose estimator~\cite{thalhammer2022cope}. The method only requires RGB images and an object model and does direct pose regression and detection in an end-to-end fashion. 

\cite{thalhammer2022cope} differs from most traditional methods dor pose estimation as it does not requires a preliminary detection stage and instead couple together the process of finding object classes and corresponding geometric correspondences, similar to~\cite{hodan2020epos}. Thus the method is also agnostic to the number of instances of the object in the scene. 

The method takes an RGB image and a 3D model as input. The initial model is a multi-scale feature pyramid network, which takes input RGB image and generates as object hypothesis, object class, geometric correspondences and 6D pose

\subsubsection{Training Data}
   
\begin{figure}[t]
      \centering
      \includegraphics[scale=0.35]{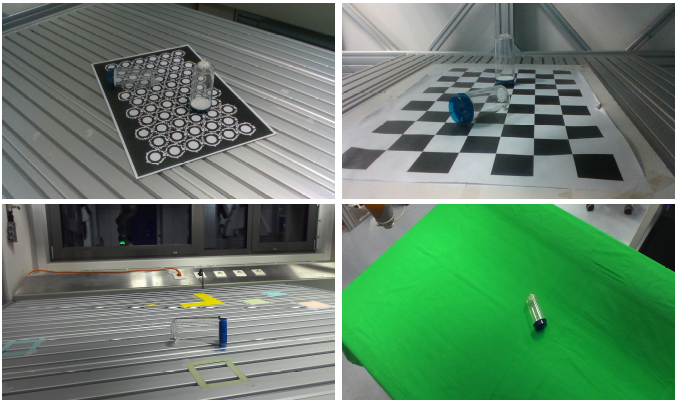}
      \caption{\textbf{Dataset} Example samples from our training dataset showing the full variation of the provided backgrounds.}
      \label{Dataset}
   \end{figure}
   
\begin{figure}[t]
      \centering
      \includegraphics[scale=0.238]{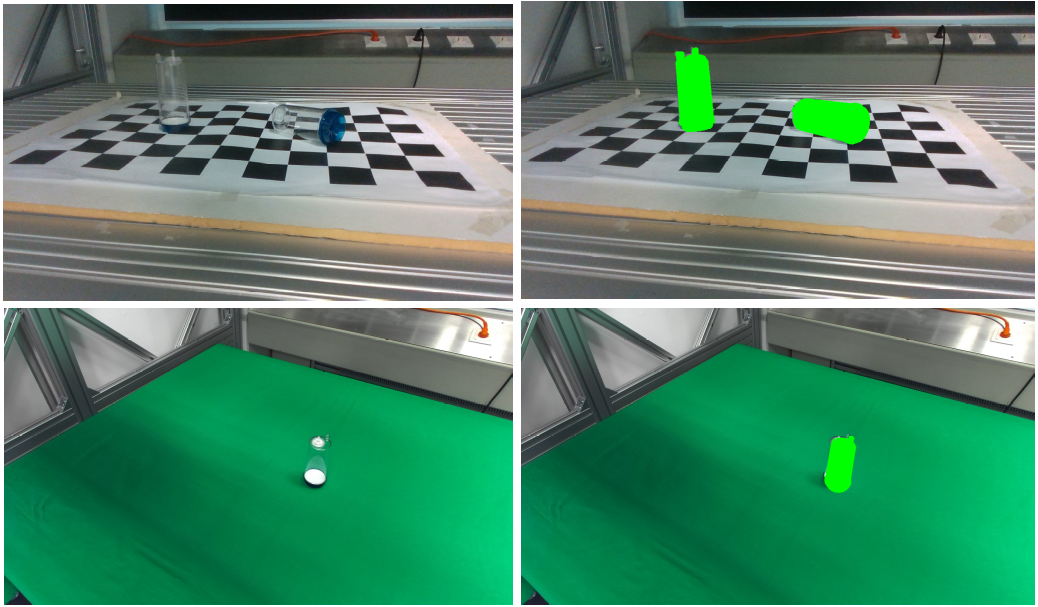}
      \caption{\textbf{Dataset annotation examples} Example annotated images from the dataset. On the left column we see the original RGB images and on the right the corresponding annotations.}
      \label{DatasetAnn}
   \end{figure}

For our data capturing process, we use the Realsense D435\footnote{\url{https://www.intelrealsense.com/depth-camera-d435/}} and the ZED from stereolabs\footnote{\url{https://www.stereolabs.com/assets/datasheets/zed2-camera-datasheet.pdf}}. 
The camera is attached to the end-effector of the KUKA arm robot\footnote{\url{https://www.kuka.com/de-at/produkte-leistungen/robotersysteme/industrieroboter/lbr-iiwa}} and moved around the object in a sequence, where for each sequence around the object 104 images are taken uniformly sampled from various heights, angles, and distances around the object in a particular pose. Since the sequences are defined manually, the pose of the camera relative to the origin of the robot is known for each captured image.

For estimating the pose of the object i.e, relative to the camera attached to the robot end-effector we leverage multi-view geometry. Information of which is provided by multiple images taken in a sequence around the object, this is used by our pose annotation tool~\cite{markus20193dsat}.

In total, we record 15 sequences each containing 104 images, of which 6 sequences were captured with only one transparent object instance and 9 with two instances of the object. This results in 1352 training images in total. Since we are dealing with transparent objects, it is even more vital for us to make our method robust to illuminations and backgrounds even more so as compared with non-transparent objects. Hence, we also use varying amounts of environment lighting and background patterns while data capturing. Particularly we use various dotted patterns, checkerboard patterns, metallic surfaces, etc for making our method robust to the background variation. For robustness to illumination, we make use of varying environment lighting and natural lights. For each sequence being captured, we vary object pose, instances, background, and illumination. Examples of our dataset can be seen in the Figure~\ref{Dataset}. 

For 6D pose annotation of the object i.e, the transformation from the camera to the object in the given scene we use 3D-SAT ~\cite{markus20193dsat}. Which is a state-of-art method for object pose annotation of RGB/RGBD sequences. One of the strongest relevance for our work is that the depth data is not necessary unlike other methods ~\cite{markus20193dsat} and thus enabling the annotation of objects that are unsuitable for depth-based methods.

The annotation tool~\cite{markus20193dsat} requires a 3D model of the object being pose annotated along with the recorded sequence of images and camera intrinsics. After recording a sequence the available data is imported to the Blender annotation GUI. This enables the alignment of 3D object models with the imported RGB images to retrieve the 6D pose of objects. Pose annotation is done by aligning the 3D model to the object in each image of the sequence. Figure ~\ref{DatasetAnn} shows a few examples of our 6D pose annotations, where the 3D model is aligned with the object. Multiple grasp poses were annotated by hand for our transparent object as shown in Figure~\ref{GraspAnnotation}.


\subsubsection{Training}


The weights of the backbone are pre-trained on ImageNet~\cite{russakovsky2015imagenet} and fine-tuned for 100 epochs using the Adam~\cite{kingma2014adam} optimizer with a learning rate of $1^{-5}$ and a batch size of 8. 
To benefit more from the pre-trained feature extractor we do not update the parameters of batch normalization and the convolution layers of the first two stages of the backbone during training.

Since the training dataset is quite limited in terms of pose variations we apply translation and zoom augmentation with up to $5\%$ each.
In order to not overfit to the training data, standard image augmentations such as brightness, contrast, blur and color changes are used. Similar to~\cite{park2019pix2pose, thalhammer2021pyrapose, sundermeyer}.

\subsection{Grasping Pipeline}

Performance of object detection and pose estimation methods often deteriorate when deployed on real-world robots ~\cite{loghmani2018recognizing}~\cite{ammirato2017dataset}~\cite{bauer2020verefine}. Thus in order to evaluate our proposed method and experimental setup, we evaluated its performance in a grasping experiment using a Toyota HSR robot ~\cite{Yamamoto2018hsr, yamamoto2019development}. 

\begin{figure}[t]
      \centering
      \framebox{{\includegraphics[scale=0.3]{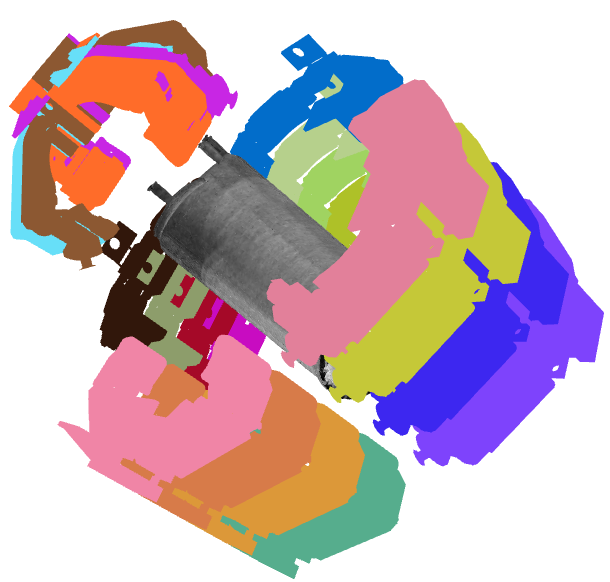}}}
      \caption{\textbf{Grasp Annotation} 20 possible grasp configurations are annotated. The randomly colored grippers are scaled to $50\%$ for visibility.}
      \label{GraspAnnotation}
   \end{figure}

Multiple grasp poses are annotated by hand for the object as shown in Figure~\ref{GraspAnnotation}. These annotated grasp poses are then transformed to the robot base frame using the estimated object pose of our method~\cite{thalhammer2022cope}. Based on the potential grasp poses, multiple trajectories are calculated and the first collision-free trajectory found is executed. A grasp is successful if the object is lifted and remains stable in the gripper.

\section{Experiments}
In this section, we present the results of our experimental setup. As mentioned previously the premise of our experimental setup is that the RGB images provide sufficient information for transparent object grasping. To evaluate the performance of our pose estimation based on our experimental setup, we evaluate it in a real-world robotic grasping experiment. 

Our grasping setup contains a Toyota HSR robot~\cite{yamamoto2019development} used for grasping our transparent canister. We place our canister on a wooden table, where the robot is looking at the canister at approximately and $45^{\circ}$ angle. We place the canister on the table in three different ways: upright position, recumbent position, and attached to the base plate. We also use two different backgrounds, in-particular we use the checkerboard and the original wooden background of our table. As the first is already part of our training dataset and the other is to see how well our method generalizes to the unseen backgrounds.

\subsection{Quantitative Results}

In this section, we describe the results of our grasping experiments. We perform in total $5$ grasps with randomized placement on the table, for the each of the four scenarios upright and recumbent with seen and unseen tabletop.
Evaluation is based on three distinct cases the grasping attempt can result in:

\begin{itemize}
\item \textit{Full Grasp}: The object is grasped and remains stable in the gripper
\item \textit{Reached Grasp}: A suitable grasp position is reached, but the grasp is unsuccessful due to the gripper moving the object previous to grasping.
\item \textit{Failed Grasp}: Neither the object is grasped nor a suitable grasping position is reached.
\end{itemize}

We assign a score of $1$, $0.5$ and $0$ for \textit{Full Grasp}, \textit{Reached Grasp} and \textit{Failed Grasp}, respectively. Reported scores in Table~\ref{table_example} are normalized by the number of grasp attempts. 










\begin{table}[t]
\caption{\textbf{Grasping Experiments} Canister grasping from a tabletop with known and unknown surface.}
\label{table_example}
\begin{center}
\begin{tabular}{|c||c|c|c|c|}
\hline

\multirow{2}{*}{Tabletop} &
      \multicolumn{2}{c|}{Seen} &
      \multicolumn{2}{c|}{Unseen} \\
    & upright & recumbent & upright & recumbent \\ \hline
\textbf{Full Grasp} & 0.6 & 0.2 & 0.4 & 0.2 \\
\hline
\textbf{Reached Grasp} & 0.1 & 0.1 & 0.1 & 0.0 \\ \hline
\textbf{overall} & 0.7 & 0.3 & 0.5 & 0.2 \\

\hline
\end{tabular}
\end{center}
\end{table}

As we see in Table~\ref{table_example}, the grasping experiments are slightly more successful in the case of the seen background as compared to the unseen background. 
Yet, the trained model generalizes to unseen object appearances.
Showing RGB provides an informative modality for transparent object handling. 
The reason for the failed grasp attempts for upright case is mainly caused by the grasping method being agnostic to object geometry. The grasping attempts have significantly deteriorated for the case of the canister being in the recumbent position, for both the seen and unseen. Although the scores remained similar for both. The reason for the significant drop in grasping performance for recumbent cases is grasp-point sampling usually picking the grasp points that are protruding the table plane, hence leading to a collision of the gripper. 

\begin{figure}[t]
      \centering
      \includegraphics[scale=0.4]{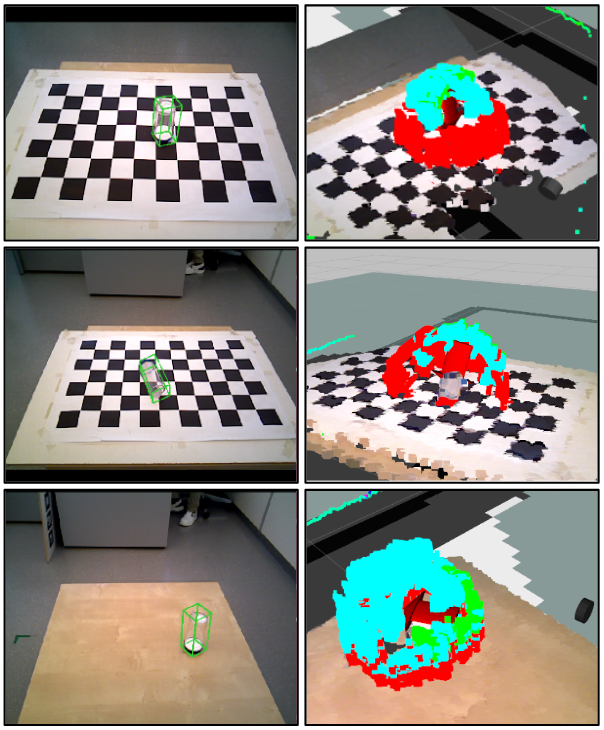}
      \caption{\textbf{Pose Estimation and Grasp Point Sampling} The left column of images indicates estimated poses with a green bounding box. Right shows all grasps, grasps protruding the table (red), grasps not protruding the table plane (blue and green), and chosen grasp (green).}
      \label{PoseEstimationGraspSampling}
   \end{figure}

\subsection{Qualitative Results}

In this section, we introduce and describe the qualitative results of our grasping experiments. We first discuss the visual results of the estimated poses and the choice of grasp points. Then we discuss the results of the grasping experiments, the cases where it succeeds, and the failure cases and give the reasoning behind them.

\subsubsection{Pose Estimation and Grasp Sampling}

The left column of Figure~\ref{PoseEstimationGraspSampling} shows the estimated 6D poses for our transparent object. We observe a small offset in the rotation of the estimated pose. The rotation is difficult to estimate since only the top part of the canister provides cues to disambiguate the rotational symmetry along the longitudinal axis of the object.

The right column of Figure~\ref{PoseEstimationGraspSampling} shows the possible grasp points around the pose estimated canister. The red points in the figure show the grasp points that are protruding the table plane, while the blue and green are the valid graspable points not protruding the table's surface. The top row shows that for the upright position of the canister it is easy to plan executable grasp trajectories. While the middle row of Figure ~\ref{PoseEstimationGraspSampling} shows the graspable points for the canister in the recumbent position.

In the case of the canister being in a recumbent position (Figure ~\ref{PoseEstimationGraspSampling} middle row), we have seen the performance drop in grasping successes, Table~\ref{table_example}. This is mainly because of the main axis of the canister lying along the table surface, resulting in the grasp point being chosen close to the table surface (red grasp points). This leads to the collision of the gripper with the surface, leading to failed grasps. 
Poses are slightly worse for the scenario featuring tabletops unseen during training time, bottom row of Figure \ref{PoseEstimationGraspSampling}.


\subsubsection{Grasping the Canister}

The top two rows of Figure ~\ref{PoseEstimationGraspSampling} show some of the examples of successful grasps. We also introduced a distractor, i.e the base plate of the canister in our grasping experiments as seen in Figure~\ref{GraspingSequenceDistractor}, which is not part of our dataset for training the pose estimator. The trained model generalizes well, with successful grasps to cases with added distractor (canister base-plate) and unseen tabletops.


\subsubsection{Failure Cases}

\begin{figure}[t]
      \centering
      \includegraphics[scale=0.37]{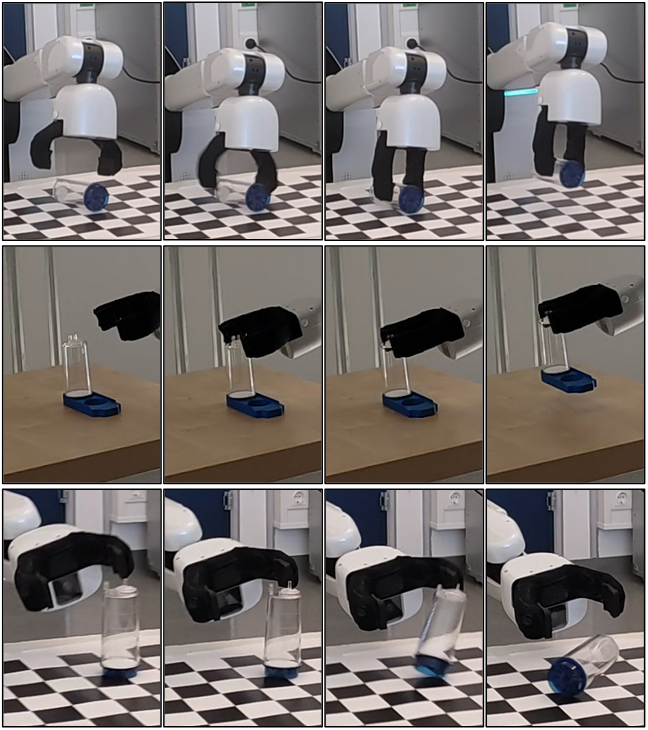}
      \caption{\textbf{Grasping Sequence with a Distractor} The Canister placed in its base plate, as such unseen during training, is picked from the table.}
      \label{GraspingSequenceDistractor}
   \end{figure}

The last row of Figure~\ref{GraspingSequenceDistractor} shows one of the failure cases for the canister in the upright position. As mentioned above, grasping sometimes fails since grasp planning does not account for the object geometry. Significant improvements can already be achieved by improving grasp point sampling and grasp trajectory planning. Additionally, providing a richer and more diverse dataset, in terms of table top texture and pose variations will improve pose estimation and thus the grasping success. 


\section{Summary and Outlook}


In this work, we conduct a study to evaluate the effectiveness of the RGB-only image space for transparent object pose estimation for robot object grasping. The experiments and the results prove our assumption about the usability of RGB
We conduct several successful grasps on a transparent object, even in an completely unseen setting, saying scene and background. 
Future work will investigate and provide improvements by training the proposed setup on a much larger scale of data including more instances and variations in the scene.
As well as improvements for grasp sampling, grasp filtering and grasp planning.

\section{Acknowledgement}

We gratefully acknowledge the support of the EU-program EC Horizon 2020 for Research and Innovation under grant agreement No. 101017089, project TraceBot, the support by the Austrian Research Promotion Agency (FFG) under grant agreement No. 879878, project K4R and the NVIDIA Corporation for the donation of the GPU used for this research.
Furthermore we would like to thank Bernhard Neuberger for supporting this work with his knowledge and expertise.

\addtolength{\textheight}{-9.0cm}


{\small
\bibliographystyle{IEEEtran}
\bibliography{bibli}
}

\end{document}